\DocumentMetadata{}
\documentclass[sigconf]{acmart}

\usepackage{booktabs} 

\setcopyright{none}

\renewcommand\footnotetextcopyrightpermission[1]{} 
\acmConference[ICVGIP'24]{15th Indian Conference on Computer Vision, Graphics and Image Processing}{December 2024}{Bangalore, India}
\acmYear{2021}
\copyrightyear{2021}

\acmPrice{15.00}

\begin{document}
\title{Learning from Limited and Imperfect Data }

\author{Harsh Rangwani }

\affiliation{%
  \institution{Adobe Research and IISc, Bangalore} 
 \state{  Young Researcher Symposium, ICVGIP 2024}
 \country{India}
  \\ \href{www.rangwani-harsh.github.io}{www.rangwani-harsh.github.io}
}

\renewcommand{\shortauthors}{}

\begin{abstract}
Deep Neural Networks have demonstrated orders of magnitude improvement in capabilities over the years after AlexNet won the ImageNet challenge in 2012. One of the major reasons for this success is the availability of large-scale, well-curated datasets. These datasets (e.g., ImageNet, MSCOCO, etc.) are often manually balanced across categories (classes) to facilitate learning of all the categories. This curation process is often expensive and requires throwing away precious annotated data to balance the frequency across classes. This is because the distribution of data in the world (e.g., internet, etc.) significantly differs from the well-curated datasets and is often over-populated with samples from common categories. The algorithms designed for well- curated datasets perform suboptimally when used to learn from imperfect datasets with long-tailed imbalances and distribution shifts. For deep models to be widely used, getting away with the costly curation process by developing robust algorithms that can learn from real-world data distribution is necessary. Toward this goal, we develop practical algorithms for Deep Neural Networks that can learn from limited and imperfect data present in the real world. These works are divided into four segments, each covering a scenario of learning from limited or imperfect data. The first part of the works focuses on Learning Generative Models for Long-Tail Data, where we mitigate the mode-collapse for tail (minority) classes and enable diverse aesthetic image generations as head (majority) classes. In the second part, we enable effective generalization on tail classes through Inductive Regularization schemes, which allow tail classes to generalize as the head classes without enforcing explicit generation of images.  In the third part, we develop algorithms for Optimizing Relevant Metrics compared to the average accuracy for learning from long-tailed data with limited annotation (semi-supervised), followed by the fourth part, which focuses on the effective domain adaptation of the model to various domains with zero to very few labeled samples.

\end{abstract}

%
%


\maketitle

\section*{Extended Abstract}

 \noindent \textbf{Generative Models for Long-Tail Data.} We first evaluate generative models' performance, specifically variants of Generative Adversarial Networks (GANs)~\cite{goodfellow2014generative} on long-tailed datasets. The GAN variants suffer from either mode-collapse or miss-class modes during generation. To mitigate this, we propose \emph{Class Balancing GAN with a Classifier in the Loop}~\cite{rangwani2021class}, which uses a classifier to asses the modes in generated images and regularizes GAN to produce all classes equally. To alleviate the dependence on the classifier, we take a closer look at model behavior at mode collapse and observe that spectral norm explosion of Batch Norm parameters correlates with mode collapse. We develop an inexpensive \emph{group Spectral Regularizer (gSR)}~\cite{hrangwani2022gsr} which mitigates the spectral collapse and significantly improves the SotA conditional GANs (SNGAN and BigGAN) performance on long-tailed data. However, we observed that class confusion is present in the generated images due to gSR norm regularization for large datasets. For this, in our latest work \emph{NoisyTwins}~\cite{rangwani2023noisytwins}, we factor the latent space as distinct Gaussian by design for each class, enforcing class consistency and intra-class diversity using a contrastive approach (BarlowTwins). This helps us to scale high-resolution StyleGANs for thousand class long-tailed datasets of ImageNet-LT and iNaturalist2019, achieving State-of-the-Art (SotA) results while maintaining class-consistency.

 \noindent \textbf{Inducting Regularization Schemes for Long-Tailed Data.} While Data Generation is promising for improving classification models on tail classes, it often comes with the cost of training an auxiliary generative model. Hence, lightweight techniques such as higher loss weights for tail classes (loss re-weighting) while training CNNs, is practical to improve performance on the minority classes. However, we observe that the model converges to saddle point instead of minima for tail classes, hindering generalization.  We show that inducing inductive bias of \emph{escaping saddles and converging to minima} for tail classes, using Sharpness Aware Minimization (SAM) significantly improves performance on tail classes~\cite{hrangwani2022escape}. Despite inductive regularizations,   training Vision Transformers (ViTs) for long-tail recognition is still challenging due to the complete lack of inductive biases such as locality of features, making their training data intensive. We propose \emph{DeiT-LT}~\cite{rangwani2024deitlt}, which introduces OOD and low-rank distillation from CNNs, to induce CNN-like robustness into scalable ViTs.

 \noindent \textbf{Semi-Supervised Learning by Optimizing Practical Metrics.} The above methods work in supervised long-tail learning, where they avoid throwing off the annotated data. However, the real benefit of long-tailed methods could be leveraged when they utilize the extensive unlabeled data present (i.e. semi-supervised setting). For this, we introduce a paradigm where we measure the performance using relevant metrics such as worst-case recall and recall H-mean on a held-out set, and we use their feedback to learn in a semi-supervised long-tailed setting. We introduce \emph{Cost-Sensitive Self Training (CSST)}~\cite{hrangwani2022cost}, which generalizes self-training based semi-supervised learning (e.g. FixMatch, etc.) to the long-tail setting with strong guarantees and empirical performance. The general trend these days is to use self-supervised pre-training to obtain a robust model and then fine-tune it. In this setup, we introduce \emph{SelMix}~\cite{ramasubramanian2023selmix}, an inexpensive fine-tuning technique to optimize the relevant metrics using pre-trained models. In SelMix, we relax the assumption that unlabeled distribution is similar to the labeled one, making models robust to distribution shifts.

 \noindent \textbf{Efficient Domain Adaptation.} The long-tail learning algorithms focus on the limited data setup and improving in-distribution generalization. However for practical usage, the model must learn from imperfect data and perform well across various domains. Towards this goal, we develop \emph{Submodular Subset Selection for Adversarial Domain Adaptation}~\cite{rangwani2021s}, which carefully selects a few samples to be labeled for maximally improving model performance in the target domain. To further improve the efficiency of the Adaptation procedure, we introduce \emph{Smooth Domain Adversarial Training (SDAT)}~\cite{hrangwani2022sdat}, which converges to generalizable smooth minima. The smooth minimum enables efficient and effective model adaptation across domains and tasks.

\section*{Future Directions}

In the above works, we propose efficient and effective methods to learn from limited and imperfect data. Based on our findings in this thesis, we present below a few prospective research directions that could be explored. 

     \noindent \textbf{Long Tail Learning in Foundational Generative Models}
    Recently, the foundation generative models like Stable Diffusion, DALLE-2, etc. have shown
    promise, with the ability to generate images based on the text description. These models are trained on internet-scale datasets using large-scale computational resources. The Internet scale datasets also follow a long-tailed data distribution due to the inherent nature of the data on the web. Hence, a good future direction is to quantify the effect of long-tailed data on text-2-image generations. Further improving the data efficiency of these models across long-tail categories is a good avenue for future.

     \noindent \textbf{Quantification of Learning from Head Classes to Tail Classes}
    One of the major focuses of long-tail learning is to be able to transfer knowledge from the populated head classes to tail classes. Despite this being the main goal, it's hard to easily quantify that how much knowledge from head classes is infused into tail classes. As for tail classes the model can either learn from head classes or learn from few-shot tail data. Hence, quantifying the learning of the tail classes in terms of head classes and few-shot learning is a necessity. Further, formalizing this notion of long-tail learning generalization bounds in terms of head-class errors and few-shot errors is a good direction to work theoretically. This makes this direction promising, having avenues of contribution.

    \noindent \textbf{Compositional Generalization for Long-Tailed (Few-Shot) Data.} One of the foundational models' features is their capability to generalize using few-shot samples. This behavior results from the model's ability to utilize the knowledge from head categories to identify tail categories with only a few samples. For example, let's say we give the model a task to identify a rare species like a blue hummingbird; in such a case, if the model has learned a generic representation (e.g., semantic parts) of a common hummingbird, it can just do some specific modifications to parts for tailoring it to a rare hummingbird. However, most theoretical works have only focused on analyzing the i.i.d. (independently and identically distributed) case. Hence, there is a requirement to re-think generalization by keeping the compositional properties in mind, particularly in the long-tailed setting.

\section*{Conclusion}
These above works explore the problem of learning neural networks from limited and imperfect data.  We have introduced the following challenges faced while training deep networks on limited and imperfect datasets: i) generative models suffer from missing modes or mode-collapse for long-tailed datasets, ii) recognition models overfit to a small set of images in tail classes, iii) overemphasis on just validation accuracy leads to poor tail class performance in long-tailed recognition and iv) distribution shift between training domain and testing domain at the time of deployment. Towards overcoming these challenges, we have introduced parts that aim towards i) training generative models for diverse and class-consistent generation even for tail classes, ii) inductive regularization schemes to ensure tail classes also behave like head classes, avoiding overfitting, iii) optimizing practical metrics like worst-case recall, etc. for robust long-tail learning and iv) efficiently adapting to target domain via minimal supervision. We then conclude by providing some avenues for future work in the above research domains.

\bibliographystyle{ACM-Reference-Format}
\bibliography{ICVGIP-Latex-Template}

\end{document}